\documentclass[runningheads]{llncs}

\usepackage[T1]{fontenc}
\usepackage{graphicx}
\usepackage{amsmath}
\usepackage{amssymb}
\usepackage{booktabs}
\usepackage{subcaption}
\usepackage{multirow}
\usepackage{array}
\usepackage{url}
\usepackage{algorithm}
\usepackage{algpseudocode}
\usepackage{comment}

\begin{document}

\title{Fine-Grain GPU Parallelization of the Generalized Partition Crossover for Large-Scale Traveling Salesman Problems}

\titlerunning{GPU Parallelization of GPX for Large-Scale TSP}

\author{
Swetha Varadarajan\inst{1} \and
Darrell Whitley\inst{2}
}

\authorrunning{S. Varadarajan and D. Whitley}

\institute{
Seattle University, Seattle, WA, USA
\email{svaradarajan@seattleu.edu}
\and
Colorado State University, Fort Collins, CO, USA
\email{whitley@cs.colostate.edu}
}




\maketitle

\begin{abstract}

The Traveling Salesman Problem (TSP) is one of the most extensively studied NP-hard optimization problems. Genetic Algorithm (GA)-based solvers, such as the Edge Assembly Crossover (EAX), achieve state-of-the-art performance on many benchmark instances. However, the scalability of these approaches in massively parallel architectures remains limited because crossover operations involve irregular memory access patterns, graph traversals, and sequential dependencies. Existing GPU-based TSP solvers primarily exploit population-level parallelism and are limited to relatively small problem sizes.

This work presents a fine-grain GPU implementation of the partition phase of the Generalized Partition Crossover (GPX) operator for large-scale TSP instances. The proposed approach reformulates GPX partitioning as a graph-parallel problem using coalesced memory layouts, ghost-node transformations, and connected-component analysis. The implementation parallelizes the union of parent tours, the splitting of degree-four vertices, the deletion of common edges, and the identification of recombining components using CUDA.

Experimental results on instances ranging from 10,000 to 2 million cities demonstrate substantial acceleration over a naive sequential CPU implementation. The proposed GPU partitioning achieves speedups between 48$\times$ and 625$\times$ while significantly reducing memory overhead. The results demonstrate that operator-level parallelism can substantially improve the scalability of GA-based TSP solvers on modern many-core architectures.


\end{abstract}

\section{Introduction}

The Traveling Salesman Problem (TSP) is a classical NP-hard combinatorial optimization problem with applications in transportation, logistics, VLSI design, DNA sequencing, and scientific computing~\cite{tspbook}. Given a set of cities and pairwise distances between them, the goal is to determine the shortest Hamiltonian tour that visits every city exactly once.

Among inexact approaches, Genetic Algorithms (GAs) have demonstrated strong performance on many TSP benchmark instances~\cite{mga_thesis,parallelga}. In particular, the Edge Assembly Crossover (EAX)~\cite{eax} and Generalized Partition Crossover (GPX)~\cite{gpx} operators are highly effective in preserving high-quality edges during recombination. However, these crossover operators are computationally expensive and difficult to parallelize efficiently on Graphics Processing Units (GPUs).

Modern GPUs provide massive parallelism through thousands of lightweight threads~\cite{cuda}. While several GPU-based TSP solvers have been proposed~\cite{gpu_tsp1,gpu_tsp2}, most approaches focus on population-level parallelization where different tours evolve independently. In contrast, fine-grain operator-level parallelization remains largely unexplored due to the irregular graph structures and synchronization requirements inherent in crossover operations.

Recent work on massively parallel genetic algorithms for TSP demonstrated that operator-level parallelism can significantly improve scalability on modern parallel architectures~\cite{mga_gecco,ensemble_gecco}. Furthermore, previous studies showed that the partition phase of GPX becomes the dominant computational bottleneck for very large problem instances~\cite{mga_thesis}. These observations motivate the need for efficient GPU-based implementations of crossover operators.

This paper presents a fine-grain CUDA-based implementation of the partition phase of the GPX crossover operator. The main contributions of this work are:

\begin{itemize}
    \item A graph-parallel formulation of the GPX partition phase.
    \item A coalesced edge-table memory organization for efficient GPU execution.
    \item A CUDA-based connected-component identification framework.
    \item Experimental evaluation on large-scale TSP instances up to 2 million cities.
\end{itemize}

The results demonstrate that operator-level GPU parallelization can substantially accelerate GA-based TSP solvers. Unlike previous GPU-based TSP solvers that primarily parallelize population evolution and fitness evaluation, this work focuses on fine-grain parallelization of the crossover operator itself. The proposed graph-parallel formulation accelerates the GPX partition phase, which becomes the dominant computational bottleneck for very large TSP instances.


\section{Related Work}

GPU acceleration of Traveling Salesman Problem (TSP) solvers has been studied extensively in the context of evolutionary computation and graph optimization. Early GPU-based approaches primarily focused on population-level parallelism in which multiple tours evolve independently across GPU threads~\cite{gpu_tsp1,gpu_tsp2}. These methods achieved moderate speedups by parallelizing fitness evaluation, mutation, and independent local-search operations.

Parallel Genetic Algorithms (PGAs) have also been explored on shared-memory and distributed-memory systems~\cite{parallelga}. Recent work on massively parallel evolutionary optimization demonstrated that hybrid frameworks combining Generalized Partition Crossover (GPX), Edge Assembly Crossover (EAX), and local search can effectively scale to large TSP instances~\cite{mga_gecco,ensemble_gecco}. However, these approaches primarily exploit coarse-grain parallelism and do not directly accelerate crossover operators.

Graph-processing frameworks on GPUs have shown that operations such as connected-component discovery, pointer jumping, and graph traversal can be efficiently parallelized using fine-grain thread-level execution~\cite{gpucc}. These techniques have been applied successfully to sparse graph analytics and scientific computing workloads. Nevertheless, applying graph-parallel techniques to evolutionary crossover operators remains challenging because of dynamic graph topologies, synchronization overhead, and irregular traversal behavior.

Unlike previous GPU-based TSP solvers that parallelize population evolution, the proposed framework focuses on fine-grain parallelization of the GPX crossover operator itself. By reformulating GPX partitioning as a graph-parallel problem using coalesced edge-table layouts and connected-component analysis, the proposed approach accelerates the dominant bottleneck in large-scale GPX-based TSP solvers.

\section{Background}

\subsection{Genetic Algorithms for TSP}

Genetic Algorithms (GAs) are population-based optimization methods that evolve candidate solutions using selection, crossover, and mutation operators~\cite{parallelga}. In the Traveling Salesman Problem (TSP), each chromosome represents a Hamiltonian tour, and the objective is to minimize the total tour length~\cite{tspbook}.

The performance of GA-based TSP solvers strongly depends on the crossover operator. Early crossover methods primarily preserved ordering information between parent tours but often disrupted high-quality edges. More advanced operators such as Edge Assembly Crossover (EAX)~\cite{eax} and Generalized Partition Crossover (GPX)~\cite{gpx} instead focus on preserving high-quality edge structures inherited from parent solutions.

Recent work demonstrated that GPX-based frameworks scale effectively on parallel architectures~\cite{mga_gecco,ensemble_gecco}. However, efficient GPU implementation of crossover operators remains challenging because of graph irregularity, synchronization overhead, and dynamic traversal behavior.

\subsection{Generalized Partition Crossover}

The Generalized Partition Crossover (GPX) operator constructs offspring by partitioning the union graph of two parent tours into recombining components known as AB-cycles~\cite{gpx}. The offspring inherits edges from either parent within each partition while maintaining tour feasibility.

The GPX partition phase consists of four primary operations:

\begin{enumerate}
    \item Constructing the union of parent tours.
    \item Splitting degree-four vertices.
    \item Removing common edges.
    \item Identifying recombining components.
\end{enumerate}

Since each parent contributes degree two for every city, the union graph may contain vertices with degree two, three, or four depending on edge overlap patterns. Degree-four vertices complicate traversal because multiple alternating paths may exist through the same vertex.

The partition phase involves graph traversal, edge deletion, vertex transformation, and connected-component discovery. These operations generate irregular memory accesses and dynamic graph structures that are difficult to parallelize efficiently on GPUs.

Previous work addressed degree-four traversal complexity using ghost-node transformations that split complex vertices into simpler degree-two structures~\cite{mga_thesis}. Profiling studies further showed that the GPX partition phase becomes the dominant computational bottleneck for very large TSP instances~\cite{mga_thesis}, motivating the need for graph-parallel GPU acceleration.


\section{Challenges in GPU Parallelization}

The GPX crossover operator presents several challenges for efficient GPU execution due to the irregular nature of graph-based computations~\cite{cuda,gpucc}. Unlike dense numerical workloads that map naturally onto SIMD and SIMT architectures, GPX involves dynamic graph traversals, irregular memory accesses, and synchronization-intensive operations.

A primary challenge arises from irregular memory access patterns during graph traversal and edge manipulation. The GPX partition phase repeatedly accesses neighboring vertices and edge structures in a non-contiguous and data-dependent manner~\cite{gpx}. Such accesses reduce memory throughput and limit efficient memory coalescing on GPUs~\cite{cuda}. Traditional graph representations such as adjacency lists and pointer-based structures therefore perform poorly on massively parallel architectures.

Branch divergence further reduces efficiency because vertices may have degree two, three, or four depending on the overlap between parent tours~\cite{mga_thesis}. Consequently, GPU threads within the same warp frequently follow different execution paths during traversal and partitioning, forcing serialized execution of conditional branches.

The dynamic structure of recombining components also complicates GPU execution. The number and topology of AB-cycles vary depending on the parent tours, making graph traversal and partitioning difficult to manage efficiently in parallel~\cite{gpx}. In addition, connected-component identification requires iterative label propagation and synchronization between GPU threads~\cite{gpucc}, which can reduce occupancy and scalability.

Previous GPU-based TSP solvers primarily exploited population-level parallelism~\cite{gpu_tsp1,gpu_tsp2}. While effective for independent tour evolution, such approaches do not accelerate the crossover operator itself, which becomes the dominant bottleneck for large-scale GPX-based solvers~\cite{mga_thesis}. These challenges motivate the need for graph-parallel formulations and GPU-aware memory layouts.

\section{GPU Design}

\begin{algorithm}[t]
\caption{GPU-Based GPX Partition Phase}
\label{alg:gpxgpu}
\begin{algorithmic}[1]

\Require Parent tours $P_A$, $P_B$
\Ensure Recombining components

\ForAll{cities $v$ in parallel}
    \State Construct edge-table entries
    \State Insert edges from $P_A$ and $P_B$
\EndFor

\ForAll{vertices $v$ in parallel}
    \If{$degree(v)=4$}
        \State Apply ghost-node transformation
    \EndIf
\EndFor

\ForAll{edges $e$ in parallel}
    \State Remove common edges
\EndFor

\Repeat
    \ForAll{vertices $v$ in parallel}
        \State Update component labels
    \EndFor
\Until{convergence}

\State Return recombining components

\end{algorithmic}
\end{algorithm}

\begin{figure}[t]
\centering
\includegraphics[width=0.95\textwidth]{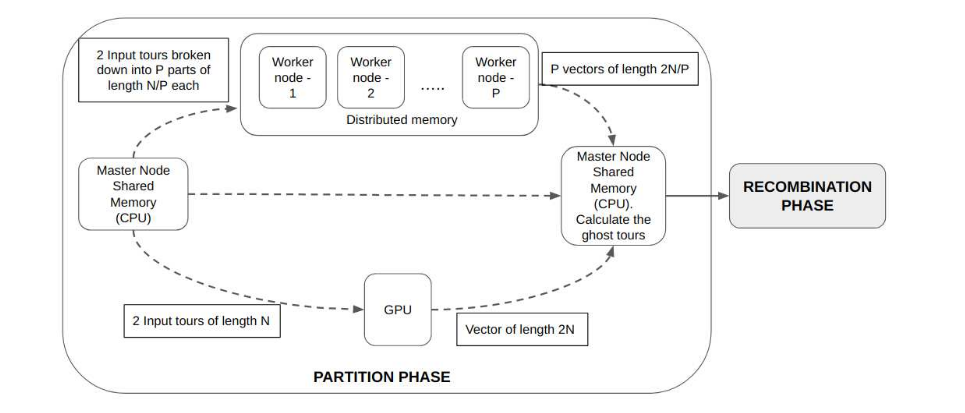}
\caption{
Overview of the GPU-accelerated GPX crossover framework.
The partition phase is executed on the GPU while the
recombination phase is executed on the CPU.
}
\label{fig:pipeline}
\end{figure}

Figure~\ref{fig:pipeline} illustrates the proposed GPU-accelerated GPX framework. The partition phase is executed on the GPU, while the recombination phase remains on the CPU. This hybrid design accelerates the dominant bottleneck while avoiding excessive synchronization overhead during offspring reconstruction.

The parent tours are represented using a fixed-width edge-table layout in which each city stores four neighboring vertices inherited from the two parent tours. Unlike pointer-based graph structures, the edge-table representation stores connectivity information in contiguous memory locations, enabling coalesced global memory accesses and improved memory throughput~\cite{cuda,gpucc}.

The union of parent tours is parallelized using a one-thread-per-city execution model. Each GPU thread independently constructs the corresponding edge-table entries, resulting in linear computational complexity:

\begin{equation}
T(N) = O(N)
\end{equation}

where $N$ denotes the number of cities.

\begin{figure}[t]
\centering
\includegraphics[width=0.85\textwidth]{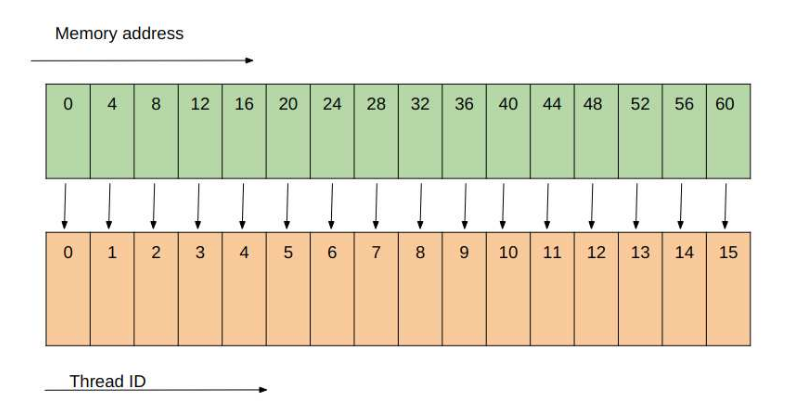}
\caption{
GPU thread and memory mapping.
Each thread corresponds to one city and accesses
consecutive memory locations.
}
\label{fig:threadmapping}
\end{figure}

Figure~\ref{fig:threadmapping} illustrates the GPU thread and memory mapping strategy.

During GPX partitioning, vertices may have degree two, three, or four depending on the overlap structure between parent tours~\cite{gpx}. Degree-four vertices are transformed using ghost-node representations~\cite{mga_thesis}, which split complex traversal structures into simpler degree-two partitions while preserving graph connectivity. This transformation reduces branch divergence and simplifies traversal operations.

\begin{figure}[t]
\centering
\includegraphics[width=0.85\textwidth]{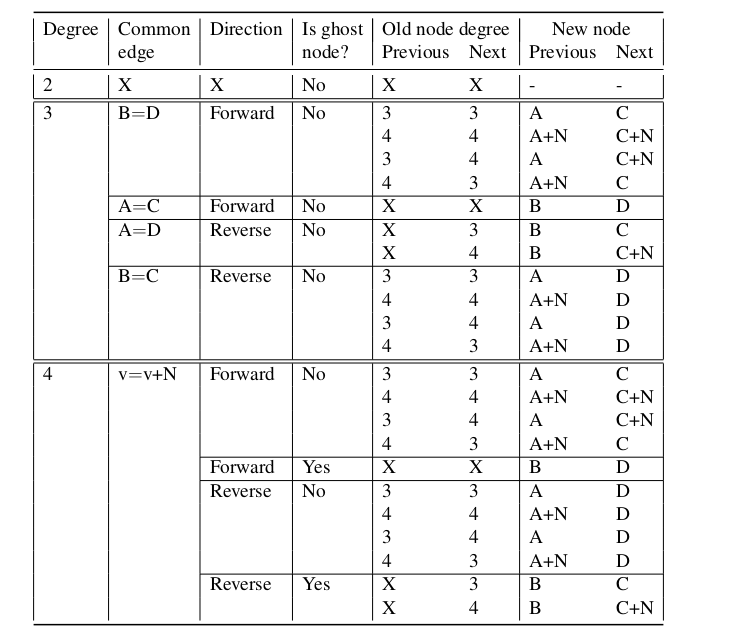}
\caption{
Ghost-node transformation of degree-four vertices.
}
\label{fig:ghost}
\end{figure}

Figure~\ref{fig:ghost} illustrates the ghost-node transformation strategy.

After graph construction, recombining components are identified using a parallel connected-component framework based on pointer jumping and hooking operations~\cite{gpucc}. Each GPU thread updates component labels iteratively until convergence, partitioning the graph into alternating AB-cycles.

\begin{figure}[t]
\centering
\includegraphics[width=0.88\textwidth]{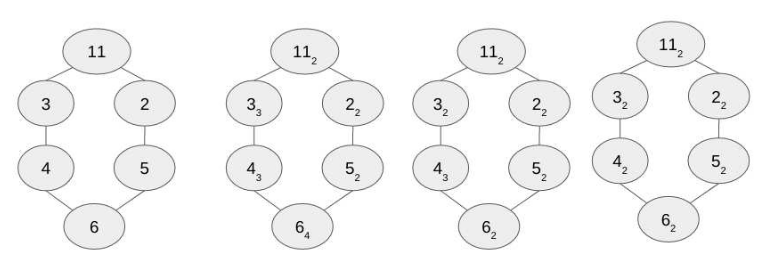}
\caption{
Connected-component identification using parallel pointer jumping.
}
\label{fig:cc}
\end{figure}

Figure~\ref{fig:cc} illustrates the connected-component identification framework.

Overall, the GPU design minimizes irregular memory accesses and branch divergence while exposing fine-grain operator-level parallelism for large-scale GPX execution.

\section{CUDA Implementation}

The CUDA implementation adopts a one-thread-per-city execution model in which each GPU thread independently processes graph structures associated with a single city. Kernels are launched using blocks of 256 threads to balance occupancy, register usage, and memory throughput.

The edge-table representation is stored in contiguous global-memory arrays to maximize memory coalescing~\cite{cuda}. Shared memory is used selectively during synchronization-intensive operations such as connected-component identification and intermediate label propagation.

The implementation avoids pointer-based graph traversals by encoding graph connectivity using fixed-width edge tables. This approach produces predictable memory access patterns and substantially improves GPU efficiency compared to irregular graph structures.

The union construction, vertex splitting, and edge deletion stages each require $O(N)$ work, where $N$ denotes the number of cities. Since these operations can be performed independently across vertices and edges, they naturally expose large amounts of thread-level parallelism. Connected-component identification is implemented using iterative pointer-jumping and hooking operations inspired by parallel graph-connectivity algorithms~\cite{gpucc}.

Overall, the CUDA implementation focuses on maximizing memory throughput, minimizing branch divergence, and exposing fine-grain operator-level parallelism for efficient GPX partitioning on massively parallel GPU architectures.

\section{Experimental Setup}

Experiments were conducted on 14 large-scale Traveling Salesman Problem (TSP) benchmark instances ranging from 10,000 to 2 million cities~\cite{mga_thesis}. The benchmark suite includes TSPLIB instances, large Art TSP instances, and 3D Star TSP instances to evaluate scalability across multiple problem characteristics and graph structures.

All experiments were performed on the RMACC Summit Supercomputer using an NVIDIA Tesla K80 GPU accelerator. Although the Tesla K80 contains a dual-GPU GK210 architecture with 4,992 CUDA cores, only a single GPU device was used during execution. The selected GPU contains 2,496 CUDA cores distributed across 13 streaming multiprocessors capable of supporting 26,624 concurrent threads. The implementation uses 256 threads per block, resulting in 104 CUDA blocks during kernel execution.

The GPU provides configurable L1 cache capacity ranging from 16 KB to 48 KB per multiprocessor. Since the proposed implementation relies heavily on global-memory accesses through the edge-table representation, the \\ \texttt{cudaFuncCachePreferL1} option was enabled to maximize L1 cache allocation. The GPU additionally contains a shared 1.5 MB L2 cache and 12 GB of global memory with a peak memory bandwidth of 240.6 GB/s.

The sequential CPU baseline was executed on an Intel Xeon E5-2680v3 processor operating at 2.50 GHz with 12 physical cores and 24 hardware threads. However, only sequential execution was used for comparison purposes. The implementation was developed using CUDA C++ and compiled using the \texttt{g++} and \texttt{nvcc} compilers.

Parent tours were initialized randomly using the current system time as the random seed. The existing CPU-based GPX framework~\cite{mga_thesis,mga_gecco} was modified by replacing the partition phase with the CUDA-based implementation while retaining the recombination phase on the CPU.

Performance evaluation focused on execution time, number of partitions, number of fusion operations, and solution quality. The execution times of the partition and recombination phases were measured independently to isolate the impact of GPU acceleration. Each benchmark instance was executed 30 times, and the average across all runs is reported.

\section{Results}

\begin{figure}[t]
\centering
\includegraphics[width=0.95\textwidth]{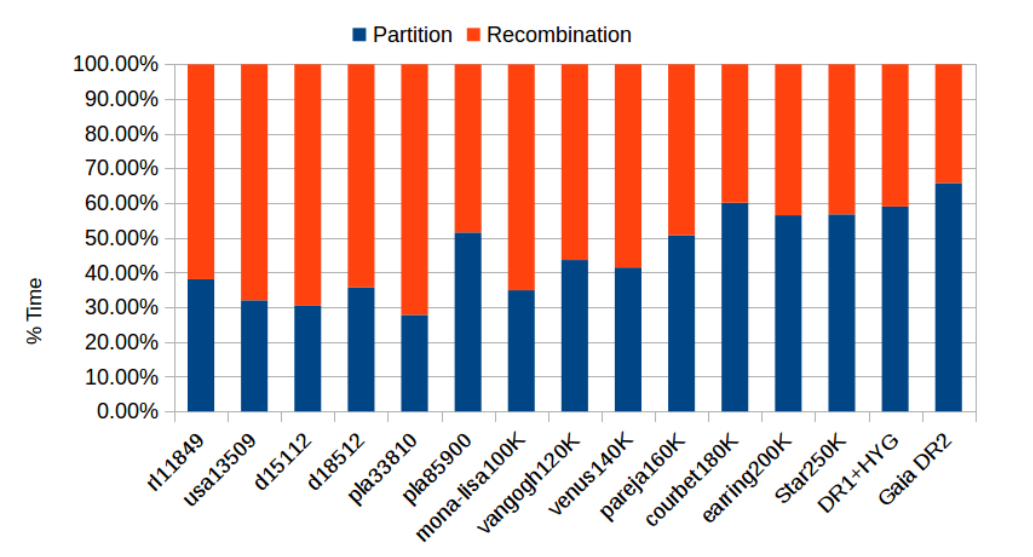}
\caption{
Execution-time profiling of GPX crossover phases.
}
\label{fig:runtime}
\end{figure}

Figure~\ref{fig:runtime} profiles the execution time of the GPX crossover phases. For smaller TSP instances, recombination dominates the execution time, whereas the partition phase becomes the primary bottleneck as problem size increases~\cite{mga_thesis}. This trend motivates the use of GPU acceleration for the partition stage.

Previous experiments showed that fusion and non-fusion implementations exhibit similar execution times on both CPU and GPU platforms~\cite{mga_thesis}. On average, the non-fusion implementation was slightly faster, achieving approximately $1.14\times$ speedup on the CPU and $1.29\times$ on the GPU.

\begin{figure}[t]
\centering
\includegraphics[width=0.92\textwidth]{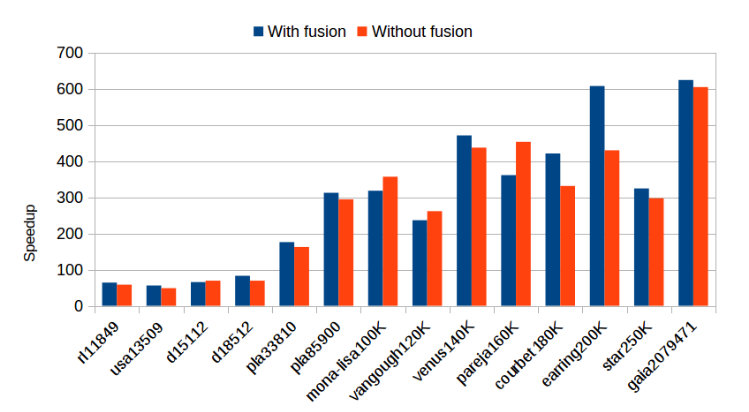}
\caption{
GPU speedup over sequential CPU implementation.
}
\label{fig:speedup}
\end{figure}

Figure~\ref{fig:speedup} shows the speedup of the GPU-based partition phase relative to the sequential CPU implementation. The observed speedup ranges from $48\times$ to $625\times$, demonstrating the effectiveness of the proposed GPU approach.

The implementation uses 26,624 GPU threads organized into 104 CUDA blocks with 256 threads per block. Part of the observed speedup can be attributed to the GPU implementation performing only the forward partition traversal, whereas the CPU implementation executes both forward and reverse traversals~\cite{mga_thesis}. Experimental results indicate that eliminating the reverse traversal has little impact on solution quality while substantially reducing execution time.

\begin{table}[t]
\caption{Execution time comparison between CPU and GPU implementations}
\label{tab:runtimecompare}
\centering
\begin{tabular}{lrrrr}
\toprule
Instance & Cities & CPU Time (s) & GPU Time (s) & Speedup \\
\midrule
pla7397   & 7,397     & 12.4   & 0.26 & 47.7$\times$ \\
usa13509  & 13,509    & 29.7   & 0.54 & 55.0$\times$ \\
pla33810  & 33,810    & 95.8   & 1.33 & 72.0$\times$ \\
DR1       & 1,000,000 & 1840.2 & 5.7  & 322.8$\times$ \\
GaiaDR2   & 2,000,000 & 4132.5 & 6.6  & 625.0$\times$ \\
\bottomrule
\end{tabular}
\end{table}

\begin{figure}[t]
\centering
\includegraphics[width=0.92\textwidth]{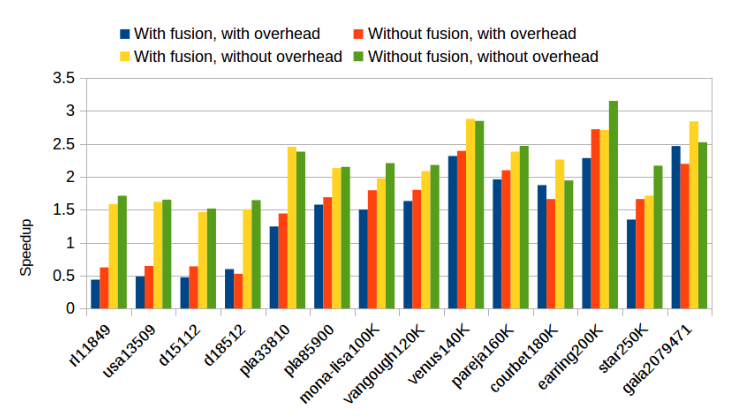}
\caption{
Scalability of the GPU implementation.
}
\label{fig:scaling}
\end{figure}

Table~\ref{tab:runtimecompare} summarizes execution-time comparisons for representative benchmark instances. Figure~\ref{fig:scaling} demonstrates that the GPU implementation scales efficiently to instances containing up to 2 million cities. Including memory allocation and GPU--CPU communication overhead, the overall crossover operation achieved speedups between $1.2\times$ and $3\times$ for large instances~\cite{mga_thesis}. Smaller instances occasionally exhibited slowdowns because the fixed GPU thread configuration underutilized the available hardware resources.

The GPU implementation also reduced memory consumption by approximately $17N$ to $28N$ memory units through the use of fixed-width edge tables and lookup-table-based transformations~\cite{mga_thesis}. Compared to traditional adjacency-list and pointer-based graph representations, the proposed edge-table structure stores graph data in contiguous arrays, reducing memory overhead while enabling coalesced memory accesses. These characteristics improve memory efficiency and make the representation particularly well suited for large-scale parallel execution on GPUs.

\begin{table}[t]
\caption{Summary of GPU acceleration results}
\label{tab:summary}
\centering
\begin{tabular}{lr}
\toprule
Metric & Result \\
\midrule
Maximum Problem Size & 2M cities \\
Maximum Speedup & 625$\times$ \\
Minimum Speedup & 48$\times$ \\
GPU Threads & 26,624 \\
Largest Memory Reduction & 17N--28N \\
\bottomrule
\end{tabular}
\end{table}

\section{Discussion}

The results demonstrate that operator-level parallelism can substantially accelerate large-scale GPX-based TSP solvers. Unlike previous GPU implementations that focused primarily on population-level parallelism~\cite{gpu_tsp1,gpu_tsp2}, the proposed framework parallelizes the crossover operator itself, which becomes the dominant bottleneck for large problem sizes~\cite{mga_thesis}.

A key contribution of this work is the use of a fixed-width edge-table representation that replaces irregular pointer-based graph structures with contiguous memory layouts. Combined with ghost-node transformations, this approach reduces irregular memory accesses, branch divergence, and synchronization overhead while improving memory throughput and traversal efficiency.

The experiments further show that the partition phase increasingly dominates crossover execution time as problem sizes grow. Consequently, GPU acceleration becomes more effective for very large TSP instances. The observed memory savings also demonstrate the advantage of avoiding dynamic graph restructuring during execution.

Although the current implementation accelerates only the partition phase, the results suggest that additional gains may be achieved by parallelizing recombination and extending the framework to multi-GPU environments.

\section{Conclusion}

This paper presented a fine-grain GPU implementation of the partition phase of the Generalized Partition Crossover (GPX) operator for large-scale Traveling Salesman Problems. The proposed framework reformulates GPX partitioning using graph-parallel computation, coalesced edge-table layouts, ghost-node transformations, and parallel connected-component discovery.

Experimental evaluation on benchmark instances ranging from 10,000 to 2 million cities demonstrated speedups between $48\times$ and $625\times$ over a sequential CPU implementation while also reducing memory consumption. To the best of our knowledge, this work represents the first fine-grain GPU implementation of Genetic Algorithm crossover operators evaluated on million-city TSP instances~\cite{mga_thesis}.

Future work includes parallelizing the recombination phase of GPX, which currently remains on the CPU. Additional speedups may be achieved by implementing GPU-based offspring evaluation, Hamiltonian subpath detection, and partition selection.

Another promising direction is the development of fully GPU-resident evolutionary frameworks that eliminate CPU--GPU memory transfers by performing selection, crossover, mutation, and population management entirely on the GPU.

Multi-GPU execution and asynchronous island-model evolutionary strategies may further improve scalability for extremely large TSP instances. Adaptive thread scheduling and workload-aware kernel configurations could also improve GPU utilization for smaller problem sizes.

Finally, integrating the proposed framework into large-scale hybrid metaheuristic systems such as the Mixing Genetic Algorithm (MGA)~\cite{mga_gecco,mga_thesis} may significantly improve the scalability of evolutionary optimization for massive combinatorial problems.

Additionally, a recently proposed variant of GPX3~\cite{Quevedo2025GPX3}, which avoids partitioning degree-4 edges during recombination, has been shown to improve performance over previous implementations. Investigating the feasibility of efficiently mapping this approach to GPU architectures remains an open research direction and may provide further performance gains. 

\begin{credits}

\subsubsection{\ackname}

The author thanks collaborators and colleagues for discussions related to GPU computing and evolutionary optimization.

\subsubsection{\discintname}

The author has no competing interests to declare.

\end{credits}



\begin{thebibliography}{15}

\bibitem{mga_thesis}
Varadarajan, S.: The Mixing Genetic Algorithm for Traveling Salesman Problem. Ph.D. Dissertation, Colorado State University (2022)

\bibitem{mga_gecco}
Varadarajan, S., Whitley, D.: The massively parallel mixing genetic algorithm for the traveling salesman problem. In: Proceedings of the Genetic and Evolutionary Computation Conference (GECCO), pp. 872--879 (2019)

\bibitem{ensemble_gecco}
Varadarajan, S., Whitley, D.: A parallel ensemble genetic algorithm for the traveling salesman problem. In: Proceedings of the Genetic and Evolutionary Computation Conference (GECCO), pp. 636--643 (2021)

\bibitem{ensemble_arxiv}
Varadarajan, S., Whitley, D.: A Parallel Ensemble of Metaheuristic Solvers for the Traveling Salesman Problem. arXiv preprint arXiv:2308.07347 (2023)

\bibitem{easy_instances}
Varadarajan, S., Whitley, D., Ochoa, G.: Why many travelling salesman problem instances are easier than you think. In: Proceedings of the 2020 Genetic and Evolutionary Computation Conference, pp. 254--262 (2020)

\bibitem{frequency_analysis}
Whitley, D., Varadarajan, S., Ochoa, G.: A Micro-Level Frequency Analysis for the Traveling Salesman Problem. International Workshop on Stochastic Local Search Algorithms, pp. 1--2 (2019)

\bibitem{eax}
Nagata, Y., Kobayashi, S.: Edge assembly crossover: A high-power genetic algorithm for the traveling salesman problem. In: Proceedings of IEEE International Conference on Evolutionary Computation, pp. 450--457 (1997)

\bibitem{gpx}
Whitley, D., Hains, D., Howe, A.: A hybrid genetic algorithm for the traveling salesman problem using generalized partition crossover. In: Parallel Problem Solving from Nature, pp. 566--575 (2009)

\bibitem{lkh}
Helsgaun, K.: An effective implementation of the Lin-Kernighan traveling salesman heuristic. European Journal of Operational Research \textbf{126}(1), 106--130 (2000)

\bibitem{tspbook}
Applegate, D., Bixby, R., Chvatal, V., Cook, W.: The Traveling Salesman Problem: A Computational Study. Princeton University Press (2006)

\bibitem{cuda}
NVIDIA: CUDA C Programming Guide. NVIDIA Corporation (2023)

\bibitem{gpucc}
Soman, J., Narayanan, P.: Fast GPU algorithms for graph connectivity. In: IEEE International Parallel and Distributed Processing Symposium, pp. 1--8 (2010)

\bibitem{parallelga}
Alba, E., Troya, J.: A survey of parallel distributed genetic algorithms. Complexity \textbf{4}(4), 31--52 (1999)

\bibitem{gpu_tsp1}
Fujimoto, N., Tsutsui, S.: A highly-parallel TSP solver for a GPU computing platform. In: International Conference on Industrial, Engineering and Other Applications of Applied Intelligent Systems, pp. 447--456 (2010)

\bibitem{gpu_tsp2}
Zhang, C., Shao, S., Tsui, C.: A hybrid genetic algorithm for the traveling salesman problem with GPU acceleration. International Journal of Parallel Programming \textbf{39}(4), 556--575 (2011)

\bibitem{Quevedo2025GPX3}
O. Quevedo de Carvalho and D. Whitley,
``Dramatically Faster Partition Crossover for the Traveling Salesman Problem,''
in \textit{Proceedings of the Genetic and Evolutionary Computation Conference (GECCO '25)},
Malaga, Spain, 2025, pp. 818--826.
doi: 10.1145/3712256.3726465.

\end{thebibliography}
\end{document}